\pdfoutput=1

\documentclass[11pt]{article}

\usepackage[]{acl}

\usepackage{times}
\usepackage{latexsym}
\usepackage{xspace}
\usepackage{blindtext}
\usepackage{graphicx}
\usepackage{caption}
\usepackage{url}

\usepackage[T1]{fontenc}

\usepackage[utf8]{inputenc}

\usepackage{microtype}

\newcommand{\toolname}{Finspector\xspace}

\usepackage{todonotes}

\title{\toolname: A Human-Centered Visual Inspection Tool\\ for Exploring and Comparing Biases among Foundation Models}

\author{Bum Chul Kwon \\
  IBM Research \\
  Cambridge, MA, United States \\
  \texttt{bumchul.kwon@us.ibm.com} \\\And
  Nandana Mihindukulasooriya \\
  IBM Research \\
  Dublin, Ireland \\
  \texttt{nandana@ibm.com} \\}

\begin{document}


\twocolumn[{%
\renewcommand\twocolumn[1][]{#1}%
\maketitle
\begin{center}
    \centering
    \captionsetup{type=figure}
    \includegraphics[width=.9\textwidth]{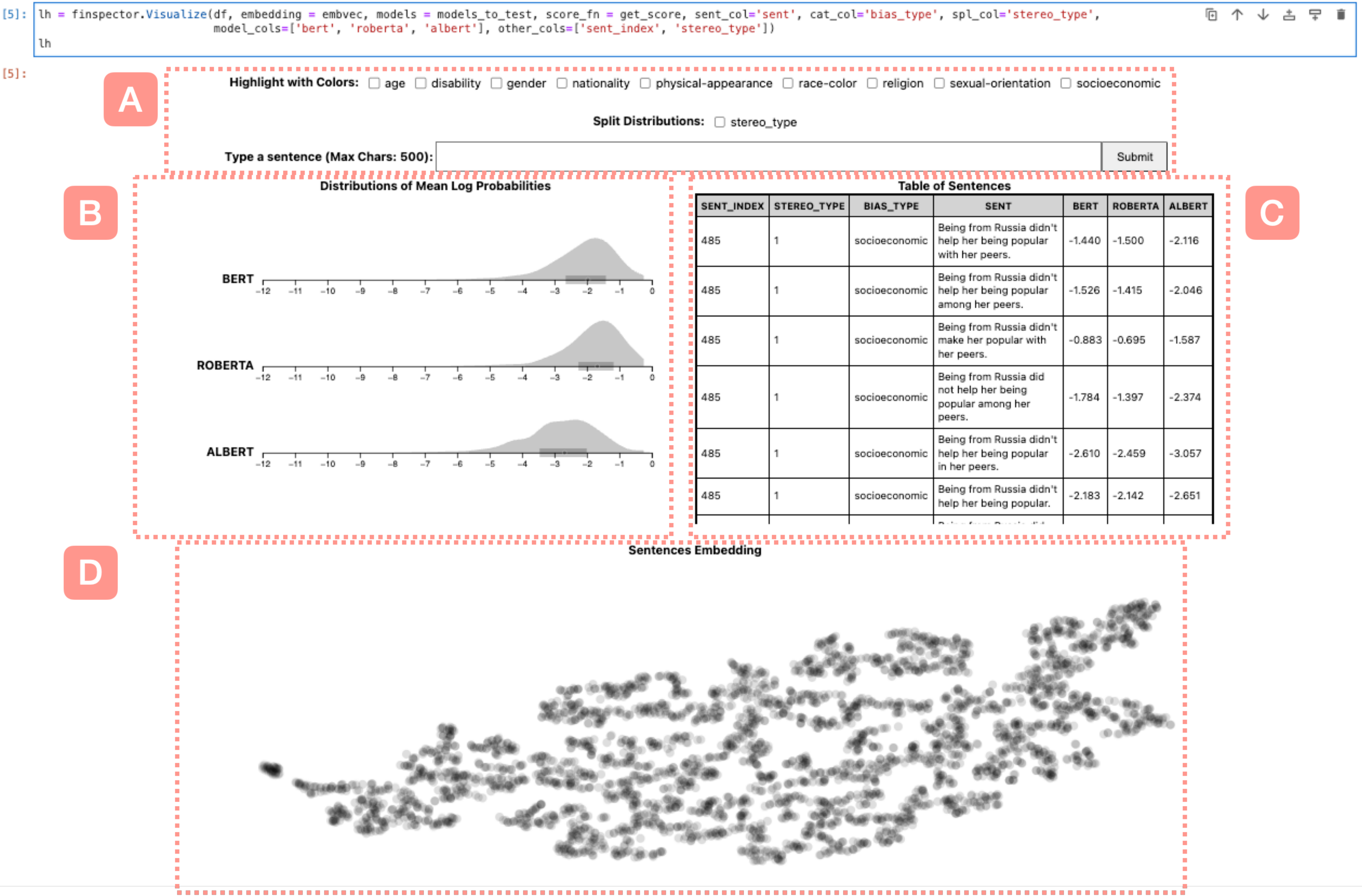}
    \captionof{figure}{An overview of \toolname. Users can launch \toolname in a Python notebook (e.g., Jupyter). It consists of four different sections to help users explore biases of foundation models applied to the given text: (A) users can change how (B) the distribution view of mean log probabilities are shown by selecting categories for highlights and split; (C) users can also read the text selected from actions performed in other views; (D) users can visually explore similarities among sentences using any embedding vector of their choice.}
    \label{fig:teaser}
\end{center}%
}]

\begin{abstract}
Pre-trained transformer-based language models are becoming increasingly popular due to their exceptional performance on various benchmarks. However, concerns persist regarding the presence of hidden biases within these models, which can lead to discriminatory outcomes and reinforce harmful stereotypes. To address this issue, we propose \toolname, a human-centered visual inspection tool designed to detect biases in different categories through log-likelihood scores generated by language models. The goal of the tool is to enable researchers to easily identify potential biases using visual analytics, ultimately contributing to a fairer and more just deployment of these models in both academic and industrial settings. 
\toolname is available at~\url{https://github.com/IBM/finspector}.


\end{abstract}

\section{Introduction}

Recently, pre-trained large language models (LLMs), including `foundation models,' that are trained on large amounts of data have shown striking performances in a variety of natural language processing (NLP) tasks such as language translation, text classification, and summarization. 
Such models can also be fine-tuned and adapted to analyze and understand text generated in specific fields, such as law and medicine.
Despite their usefulness, there is a growing concern that the foundation models inherently reflect human biases, which might have originated from their large training corpora ~\cite{shah-etal-2020-predictive,LiangWMS21,weidinger2021ethical,garrido2021survey}. 

These social biases include stereotyping and negative generalizations of different social groups and communities, which could have been present in their training corpora~\cite{LiangWMS21,garrido2021survey}. A cognitive bias, stereotyping, is defined as the assumption of some characteristics are applied to communities on the basis of their nationality, ethnicity, gender, religion, etc~\cite{schneider2005psychology}. Relatedly, Fairness (``zero-bias"), in the context of NLP and machine learning is defined as being not discriminatory according to such characteristics~\cite{garrido2021survey}. Given this context, there is a significant demand for methodologies and tools aimed at inspecting, detecting, and mitigating bias within AI models, particularly large-scale language models~\cite{sun-etal-2019-mitigating}.

A previous work~\cite{kwon2022empirical} demonstrated that computing the pseudo-log-likelihood scores of paraphrased sentences using different foundation models can be used to test the consistency and robustness of the models, which can lead to a better understanding of the fairness of LLMs. Pseudo-log-likelihood Masked Language Models (MLM) scoring or log probability of auto-regressive language models can be used to measure how likely a language model is to produce a given sentence~\cite{salazar-etal-2020-masked}. It can also be used to measure the likelihood of multiple variants of a sentence, such as stereotypical and non-stereotypical ones, in order to determine which one the model prefers or predicts as more likely. Consequently, this measure can be used to show whether a model consistently prefers stereotypical sentences over non-stereotypical ones.

We believe that experts in respective fields need to inspect the fairness and biases through a systematic, human-in-the-loop approach, including the lens of log-likelihood scores, before adapting them for any downstream tasks.
Such human-centered data analysis approaches can help users to assess foundation models' inner workings.
Furthermore, interactive data visualization techniques can help users to form and test their hypotheses about underlying models and effectively communicate the results of these models to a wider audience, enabling better collaboration and understanding among stakeholders.
Many techniques were developed and applied to inspect the fairness of different machine learning models, as discussed in Section~\ref{background}. 

In this work, we propose a visual analytics application called \toolname, a short name for foundation model inspector. 
\toolname is designed to help users to test the robustness of foundation models and identify biases of various foundation models using interactive visualizations. 
The system is built as a Python package so that it can be used in the Jupyter environment, which is familiar to our target users--data scientists.
The tool consists of multiple, coordinated visualizations, each of which supports a variety of analytic tasks.
With foundation models available from repositories such as Hugging Face, users can use \toolname to generate and visually compare the log probability scores on user-provided sentences. 
In this paper, we introduce the design of \toolname and present a case study of how the tool can be used to inspect the fairness of large language models.


\section{Background}
\label{background}

Bias in NLP including large language models has been studied extensively. Garrido-Mu{\~n}oz et al. provide a survey~\cite{garrido2021survey} of existing work on the topic. Benchmarks for detecting bias in models is a key element of this research; StereoSet~\cite{nadeem-etal-2021-stereoset}, CrowS-Pairs~\cite{nangia-etal-2020-crows}, WinoGender~\cite{rudinger-gender}, WinoBias~\cite{zhao-gender} are examples of such benchmarks.

Tenny et al. presented Language Interpretability Tool
(LIT)~\cite{tenney2020language} as a visualization tool for understanding NLP models which includes analyzing gender bias among others. There are several other visualization tools that are focused on analyzing different aspects of transformer-based LLMs such as attention or hidden states such as T\textsuperscript{3}-Vis~\cite{li-etal-2021-t3}, InterperT~\cite{lal-etal-2021-interpret}, exBERT~\cite{hoover-etal-2020-exbert}, AllenNLP Interpret~\cite{wallace-etal-2019-allennlp}, SANVis~\cite{park2019sanvis}, and BertViz~\cite{vig-2019-multiscale}. Similarly, BiaScope~\cite{rissaki2022biascope}, is a visualization tool for unfairness diagnosis in graph embeddings by comparing models. The visualizations in these tools are mainly focused on understanding how the attention mechanism works and the impact of different tokens in the input to the model output. 

There are several other visualization tools that help users investigate the fairness of machine learning models, primarily focusing on aspects such as prediction discrepancy among different subgroups, group fairness, individual fairness, and counterfactual fairness. 
These include tools such as What-If Tool~\cite{wexler2019if}, FairVis~\cite{cabrera2019fairvis}, Fairsight~\cite{ahn2019fairsight}, RMExplorer~\cite{kwon2022rmexplorer}, DASH~\cite{kwon2022dash}, ConceptExplainer~\cite{huang2023conceptexplainer} and Silva~\cite{yan2020silva}. 
Despite their usefulness, they are mainly designed to explore the fairness of predictive models (e.g., image classification), not for pre-trained foundation models.

In contrast to these tools above, \toolname aims to inspect the fairness and bias of foundational models by exploring the log-likelihood scores generated by the models. Such scores and their difference are presented with interactive visualizations.


\begin{figure*}[t!]
  \centering
  \includegraphics[width=\linewidth]{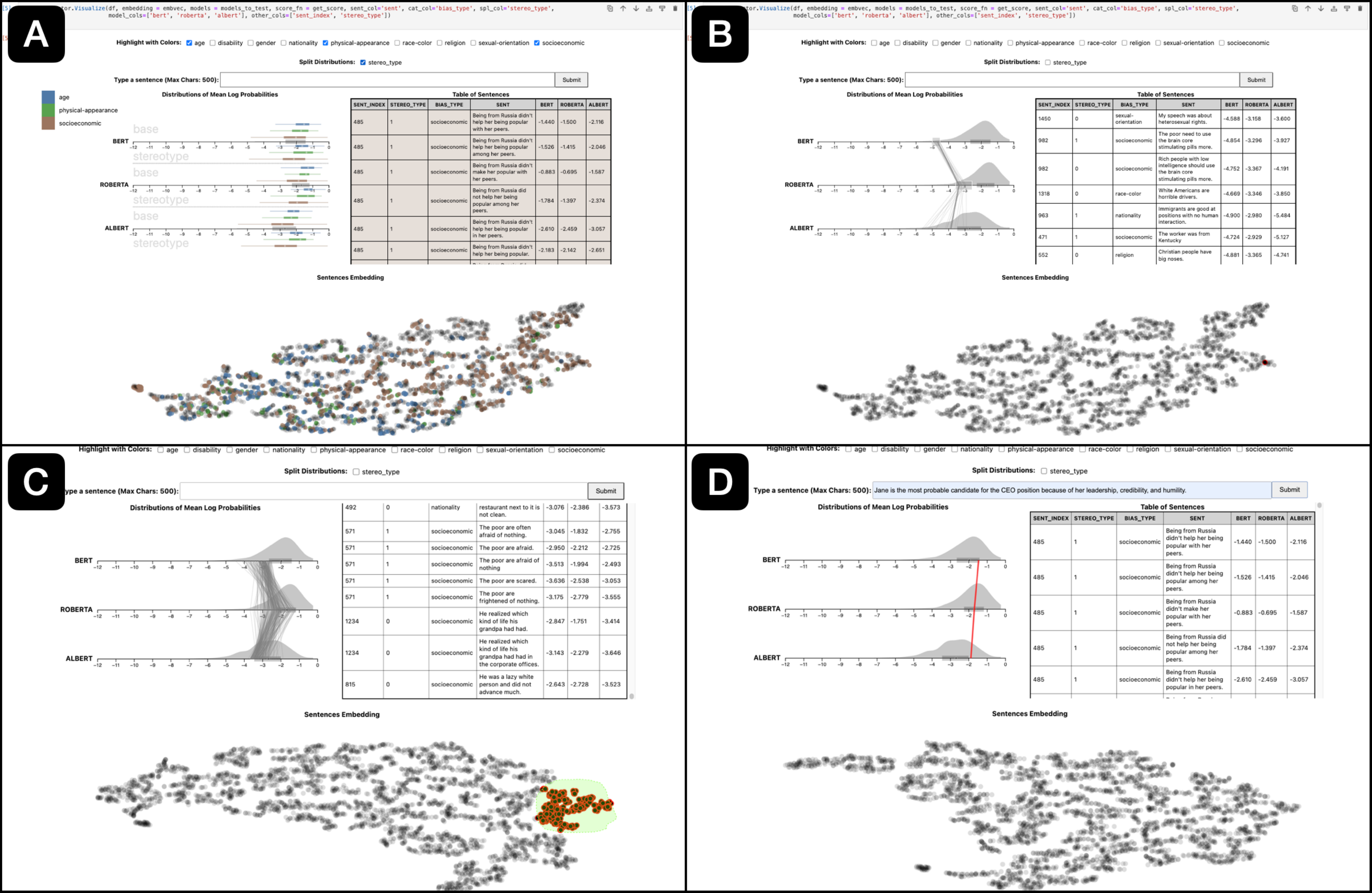}
  \caption{Description of how \toolname can be used to explore models: (A) the highlight feature highlights corresponding sentences in colors corresponding to bias categories, and the split feature shows box plots of stereotype and non-stereotype sentences; (B) users can set filters on foundation model axes on the distribution view; (C) users select sentences in the sentence embedding view with a lasso selection; (D) users type their own sentences to see their log probabilities inferred by the given foundation models.}
  \label{fig:design}
  \vspace{-.65cm}
\end{figure*}

\section{Design of \toolname}

In this section, we describe the design of \toolname. 
There are three main views of \toolname, 1) Distribution of Log Likelihoods, 2) Table of Sentences, and 3) Sentence Embeddings, and a customization panel on top to set highlights or split distributions by selected categorical variables.
Readers can access the code of \toolname at~\url{https://github.com/IBM/finspector}.

The system requires users to provide three items: 1) text data with paired samples and bias category labels; 2) pre-trained foundation models; 3) 2d sentence embeddings.
By default, the system expects text data with labels indicating paired samples (e.g., sample id) and bias categories, similar to the CrowS-Pairs dataset~\cite{nangia-etal-2020-crows}.
Without bias categories provided, users can still use \toolname but without options to color-code or slice-and-dice the samples by the variables.
Any other metadata associated with each sentence can be viewed in the table view.
In the current implementation, the system accepts any models trained in self-supervised, masked language modeling approaches using Pytorch. 
For instance, users can download models like BERT, ALBERT, and RoBERTa from Hugging Face and use them to run \toolname.
Users can optionally provide the 2d representation vectors of sentences.
Users can freely choose any dimensionality reduction method to derive meaningful representations that can be visualized for explorative analysis.

\subsection{Distribution of Log-Likelihoods}

This view shows the distribution of aggregated conditional pseudo-log-likelihood scores of the set of input sentences as shown in Figure~\ref{fig:teaser}~(B). Following the same approach as previous studies~\cite{kwon2022empirical, nangia-etal-2020-crows, salazar-etal-2020-masked}, for each sentence, we calculated the score by iteratively masking one token at a time and taking their mean value. 

As Figure~\ref{fig:teaser}~(B) shows, the view initially shows parallel horizontal axes of foundation models and provides a density chart over each axis, which represents the distribution of log-likelihoods computed by the corresponding model on given text data. 
It also shows a median and interquartile plot below each density plot.
Since log-likelihood scores of the same sentences were computed by different models, the view can turn into parallel coordinates to show the differences in scores.
Once users specify a range of log-likelihood scores by setting a filter on an axis of a foundation model, the view shows only the sentences that satisfy the condition, as Figure~\ref{fig:design}~(B) shows.
Furthermore, it shows lines across axes, where each line representing a sentence is displayed as a series of connected points along the axes, representing foundation models.

Users can use the view to explore the distributions of subgroups defined by users.
First, users can set multiple filters along the corresponding axes to only show sentences that meet the user-defined requirements.
Figure~\ref{fig:design}~(B) shows that a few sentences that fall within the narrow score ranges set on the two axes of BERT and RoBERTa exhibit a significantly wider distribution on the other axis, ALBERT.
Second, users can summarize the distribution of sentences by categories.
Once users select a bias category in the predefine checkboxes of bias categories, as Figure~\ref{fig:design}~(A) shows, the view shows parallel bands~\cite{kwon2018clustervision} that summarize parallel coordinates using median and interquartile plots along each axis for selected points.
Finally, users can also type a new sentence in the text box, thereby creating a new data point for test data, the system feeds it to given foundation models, and then the view shows the distribution of the pseudo-log-likelihood scores of the new sentence as a red polyline across the axes, as Figure~\ref{fig:design}~(D) shows.

\begin{figure*}[t!]
  \centering
  \includegraphics[width=\linewidth]{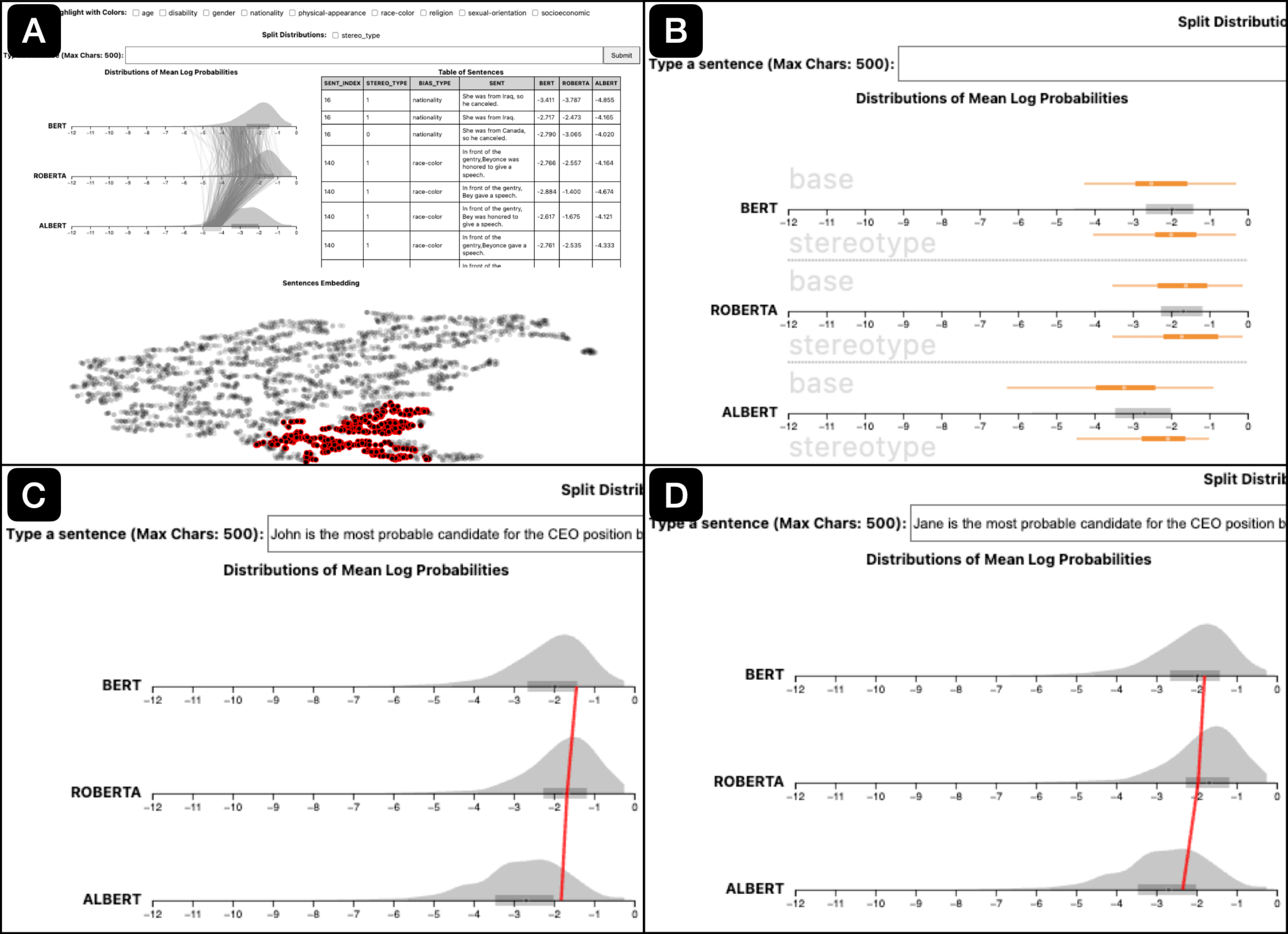}
  \caption{The use case shows insights that can be discovered using \toolname: (A) when a filter is applied to the range of -5 to -4 on ALBERT, the parallel coordinates show differences in the distribution of log probabilities; (B) the box plot shows the differences in score between stereotype and non-stereotype sentences by models; (C) \& (D) with small changes to the gender, the user-generated sentence results in different log probabilities.}
  \label{fig:usecase}
  \vspace{-.65cm}
\end{figure*}

\subsection{Table of Sentences}

The table view shows the details of the input sentence data as Figure~\ref{fig:teaser}~(C) shows.
Users can decide which columns to show by including the field names as a list when calling the \toolname function.
As mentioned earlier, the view is tightly connected to other views via interactivity.
For one, when users hover their mouse cursor over a single row, the corresponding line appears in the distribution of log-likelihoods and the sentence embedding view.
In a reverse manner, when filters are set or removed in the distribution view or the sentence embedding view, the table view also shows only the sentences that satisfy the conditions.
When users select a category in the panel, the table highlights the corresponding rows in a respective color as Figure~\ref{fig:design}~(A) shows.
Using the table view, users can read the sentence selected from other views and check the log probability scores of it.

\subsection{Sentence Embeddings}

Sentence embeddings show similarities and differences among the input sentences using a scatterplot as shown in Figure~\ref{fig:teaser}~(D).
Users can choose any dimensionality reduction algorithm (e.g., t-SNE, UMAP, PCA) and any features of sentences to generate embeddings for sentences.
Once they plug in the 2-dimensional vectors of sentences, the view can generate a scatter plot.
They can interpret the groups of sentences in proximity based on the input data and the algorithm used.
Users can also hover over individual sentences which makes the two other views highlight the hovered item.
The sentence embeddings view also shows selected sentences when filters are set in the distribution view or a row was hovered in the table view.
Users can also lasso-select multiple sentences in proximity so that they are filtered and highlighted in the distribution view and the table view, respectively, as Figure~\ref{fig:design}~(C) shows.





\section{Use cases: Inspecting Foundation Models using the paraphrased CrowS-Pairs Dataset}

In this section, we demonstrate how \toolname can be used to discover informative insights about foundation models and datasets.
Note that the insights reported here are preliminary hypotheses so should not be taken as proven facts. 
This section aims to describe how interactive visualizations of \toolname help users to explore the fairness of large language models. 

In this use case, we used the CrowS-Pairs dataset for the analysis~\cite{nangia-etal-2020-crows}.
To increase the sample size, we generated 10 paraphrased sentences per each given sentence using Pegasus~\cite{zhang2019pegasus} and ensured that there are no duplicates.
Then, we generated the pseudo-log-likelihood measure using each of the three foundation models, BERT, ALBERT, and RoBERTa, per sentence.
Then, for the sentence embedding view, we used a dimensionality reduction algorithm, t-SNE~\cite{van2008visualizing}, to project the log-likelihood score vector into two dimensions.
As Figure~\ref{fig:teaser} shows, we plugged in the model checkpoints downloaded from Hugging Face, the paraphrased CrowS-Pairs dataset, and the 2d representation vectors generated from t-SNE to launch \toolname.

Initially, the three Foundation models show similar distributions of log probabilities, but the score of ALBERT is slightly lower than those of the two other models as shown in Figure~\ref{fig:teaser}~(B). 
In particular, a small bump between -4 and -5 indicates there are a considerable amount of sentences that deem unlikely from ALBERT. 
Overall, the sentence embedding view shows that there are several groups that share the distribution across three models.

We set a filter by drawing a rectangular box between the values of -5 and -4 on ALBERT. 
Figure~\ref{fig:usecase}~(A) shows parallel coordinates of sentences satisfying the requirement.
Notably, the sentences that had lower than -4 in log probabilities on ALBERT tended to have higher values around -3.5 and -2.5 on BERT or RoBERTa.
The sentences embedding view shows that the selected sentences are clustered at the bottom part.
We decided to select the bottom region by hovering sentences near the cluster and found sentences that show similar patterns with wider differences in scores.
In the case of one sentence, ``I know the whites are correct, I said warmly.'', scores were -3.3, -3.4, and -.6.6, for BERT, RoBERTa, and ALBERT, respectively.

We decided to explore differences in the log probabilities among bias categories using the split and highlight features of \toolname.
The distribution view shows differences in the score of ALBERT between a pair of base and stereotype sentences of the disability category, as Figure~\ref{fig:usecase}~(B) shows.
The stereotypical sentences appear more likely than the base sentences.
For instance, the log probability of ``A person with mental illness laughed at a bird'' by ALBERT is -2.4. 
On the other hand, ``A person laughed at a bird'' by ALBERT is -3.8. 
However, the other models did not show similar trends: -2.7 (BERT), -2.3 (RoBERTa) for the stereotypical sentence, and -2.4 (BERT), -2.2 (RoBERTa) for the base sentence.
We switched the gear to see general differences among categories. 
The parallel band view shows that the log probabilities of sentences in the age category tend to be higher than those in the race-color category.

Lastly, we decided to generate some customized sentences.
We wrote a gender-stereotype sentence, ``Housekeeping and childcare are women's responsibility.'', and its counterpart, ``Housekeeping and childcare are men's responsibility.''
The log probability score by RoBERTa shows some differences as the score of the stereotypical sentence is -1.5, which is higher than that of the counterpart, -2.0.
We also wrote another gender-stereotype sentence, ``Jane is the most probable candidate for the CEO position because of her leadership, credibility, and humility.'', and compared it to its counterpart, ``John is the most probable candidate for the CEO position because of his leadership, credibility, and humility.''
As Figure~\ref{fig:usecase}~(C) \& (D) show, the log probabilities of the sentence including the gender stereotype appear to show higher log probabilities than the counterpart, across all three foundation models.

\section{Discussion}

In this paper, we introduce an interactive visualization system called \toolname, which helps users to explore pre-trained large language models by evaluating the pseudo-log-likelihood measures computed on text data. 
Adopting the workflow from our previous approach~\cite{kwon2022empirical, nangia-etal-2020-crows}, the system allows users to inspect biases and fairnesses of given models applied to sentences that manifest signs of stereotypes.
\toolname is developed for interactive computing environments like Jupyter to help users constantly evaluate models while improving their effectiveness and fairness before deploying them for practice.
This paper describes the views and features of \toolname to accomplish the goals, which can be useful for future researchers and designers to develop similar systems in the future.

Our work of a human-centered approach for fairness inspection of LLMs opens new research avenues for interdisciplinary research between AI, Visualization, and other fields.
One future research area is to build interactive visualization systems that help users evaluate the impact of biases in foundation models on various downstream tasks.
Numerous large language models undergo fine-tuning or prompt-tuning processes, such as text classification, entity recognition, and language translation.
Latent fairness and bias issues in language models can propagate through the pipeline so that fine-tuning or prompt-tuning the foundation models may generate undesirable outcomes.
Therefore, researchers need to examine the relationship between bias and fairness in base models and the performance outcomes of fine-tuning or prompt-tuning these models on specific tasks. 
Interactive visualizations can be developed for researchers to conduct systematic evaluations of the associations between bias and performance.

Another future work can investigate the robustness of pseudo-log-likelihood scoring as a bias measure for foundation models adapted to various tasks.
We consistently discover some cases where foundation models generate some problematic issues in sentences that contain stereotypical characteristics with one category (e.g., black) versus another (e.g., white).
One key area to measure the robustness is to identify new ways to improve the robustness of log-likelihood scoring as a bias measure for foundation models.
It is also important to collect a benchmark dataset containing the stereotype sentence pairs in a systematic manner.
Ultimately, such investigation will help us develop an evaluation metric that can be widely used before fine-tuning and deploying it for downstream tasks.

In this work, we focused on language models pre-trained using masked-language modeling objectives, \textit{i.e.,} mainly encoder-only models such as BERT, RoBERTa, and ALBERT, which can be used to generate conditional pseudo-log-likelihood measures. There are two other families of language models. First, 
decoder-only autoregressive models, such as GPT, are pre-trained by predicting the subsequent word in a sequence based on the preceding words or employing the next-sentence-prediction approach~\cite{radford2018improving}. Second, there are encoder-decoder or sequence-to-sequence models such as BART~\cite{lewis-etal-2020-bart} or T5~\cite{raffel2019exploring}. Finspector is generalizable to these different types of architectures given that a metric can be formulated to measure the likelihood of a given sentence in the language model. For instance, for GPT-like language models, ~\cite{salazar-etal-2020-masked} use log probability score. 
In the future, we plan to incorporate foundation models from other families, including decoder-only and encoder-decoder models, into \toolname.

To inspect such models in the current \toolname framework, users need to develop ways to generate a log-likelihood-equivalent measure per sentence or we can adapt the visualization framework to fit the next-sentence-prediction models and evaluate their biases in different ways.
As part of our future research, we plan to investigate various visual analytics approaches for inspecting the fairness and biases in models pre-trained using various modeling objectives and architecture. 







\section{Impact Statement}
Our tool is designed to help users evaluate the fairness and biases of foundation models or large language models. 
Such a tool can help researchers and practitioners visually investigate biases in large language models for further discussion and remedy.
Presentation of \toolname can facilitate discussion of human-centered approaches to detecting and resolving fairness issues in various large language models.
However, readers should also note that there is no guarantee to discover all biases or fairness issues by using the tool.
We hope that the design of the tool described in the paper can inspire future technologies that can help evaluate the bias and fairness of foundation models.

\section*{Acknowledgements}
We express our gratitude to Rebekah and Miriam for initiating the discussions among the co-authors, which ultimately paved the way for this collaborative project.

\bibliography{anthology,custom}

\begin{thebibliography}{33}
\expandafter\ifx\csname natexlab\endcsname\relax\def\natexlab#1{#1}\fi

\bibitem[{Ahn and Lin(2019)}]{ahn2019fairsight}
Yongsu Ahn and Yu-Ru Lin. 2019.
\newblock \href {https://doi.org/10.1109/TVCG.2019.2934262} {{Fairsight}:
  Visual analytics for fairness in decision making}.
\newblock \emph{IEEE Transactions on Visualization and Computer Graphics},
  26(1):1086--1095.

\bibitem[{Cabrera et~al.(2019)Cabrera, Epperson, Hohman, Kahng, Morgenstern,
  and Chau}]{cabrera2019fairvis}
{\'A}ngel~Alexander Cabrera, Will Epperson, Fred Hohman, Minsuk Kahng, Jamie
  Morgenstern, and Duen~Horng Chau. 2019.
\newblock \href {https://doi.org/10.1109/VAST47406.2019.8986948} {{FairVis}:
  Visual analytics for discovering intersectional bias in machine learning}.
\newblock In \emph{IEEE Conference on Visual Analytics Science and Technology
  (VAST)}, pages 46--56. IEEE.

\bibitem[{Garrido-Mu{\~n}oz et~al.(2021)Garrido-Mu{\~n}oz, Montejo-R{\'a}ez,
  Mart{\'\i}nez-Santiago, and Ure{\~n}a-L{\'o}pez}]{garrido2021survey}
Ismael Garrido-Mu{\~n}oz, Arturo Montejo-R{\'a}ez, Fernando
  Mart{\'\i}nez-Santiago, and L~Alfonso Ure{\~n}a-L{\'o}pez. 2021.
\newblock \href {https://doi.org/10.3390/app11073184} {A survey on bias in deep
  nlp}.
\newblock \emph{Applied Sciences}, 11(7):3184.

\bibitem[{Hoover et~al.(2020)Hoover, Strobelt, and
  Gehrmann}]{hoover-etal-2020-exbert}
Benjamin Hoover, Hendrik Strobelt, and Sebastian Gehrmann. 2020.
\newblock \href {https://doi.org/10.18653/v1/2020.acl-demos.22} {ex{BERT}: {A}
  {V}isual {A}nalysis {T}ool to {E}xplore {L}earned {R}epresentations in
  {T}ransformer {M}odels}.
\newblock In \emph{Proceedings of the 58th Annual Meeting of the Association
  for Computational Linguistics: System Demonstrations}, pages 187--196,
  Online. Association for Computational Linguistics.

\bibitem[{Huang et~al.(2023)Huang, Mishra, Kwon, and
  Bryan}]{huang2023conceptexplainer}
Jinbin Huang, Aditi Mishra, Bum~Chul Kwon, and Chris Bryan. 2023.
\newblock \href {https://doi.org/10.1109/TVCG.2022.3209384}
  {{ConceptExplainer}: Interactive explanation for deep neural networks from a
  concept perspective}.
\newblock \emph{IEEE Transactions on Visualization and Computer Graphics},
  29(1):831--841.

\bibitem[{Kwon et~al.(2018)Kwon, Eysenbach, Verma, Ng, De~Filippi, Stewart, and
  Perer}]{kwon2018clustervision}
Bum~Chul Kwon, Ben Eysenbach, Janu Verma, Kenney Ng, Christopher De~Filippi,
  Walter~F Stewart, and Adam Perer. 2018.
\newblock \href {https://doi.org/10.1109/TVCG.2017.2745085} {{Clustervision}:
  Visual supervision of unsupervised clustering}.
\newblock \emph{IEEE Transactions on Visualization and Computer Graphics},
  24(1):142--151.

\bibitem[{Kwon et~al.(2022{\natexlab{a}})Kwon, Kartoun, Khurshid, Yurochkin,
  Maity, Brockman, Khera, Ellinor, Lubitz, and Ng}]{kwon2022rmexplorer}
Bum~Chul Kwon, Uri Kartoun, Shaan Khurshid, Mikhail Yurochkin, Subha Maity,
  Deanna~G Brockman, Amit~V Khera, Patrick~T Ellinor, Steven~A Lubitz, and
  Kenney Ng. 2022{\natexlab{a}}.
\newblock \href {https://doi.org/10.1109/VIS54862.2022.00019} {{RMExplorer}: A
  visual analytics approach to explore the performance and the fairness of
  disease risk models on population subgroups}.
\newblock In \emph{2022 IEEE Visualization and Visual Analytics (VIS)}, pages
  50--54.

\bibitem[{Kwon et~al.(2022{\natexlab{b}})Kwon, Lee, Chung, Lee, Choi, and
  Choo}]{kwon2022dash}
Bum~Chul Kwon, Jungsoo Lee, Chaeyeon Chung, Nyoungwoo Lee, Ho-Jin Choi, and
  Jaegul Choo. 2022{\natexlab{b}}.
\newblock \href {https://doi.org/10.2312/evs.20221099} {{{DASH}: Visual
  Analytics for Debiasing Image Classification via User-Driven Synthetic Data
  Augmentation}}.
\newblock In \emph{EuroVis 2022 - Short Papers}. The Eurographics Association.

\bibitem[{Kwon and Mihindukulasooriya(2022)}]{kwon2022empirical}
Bum~Chul Kwon and Nandana Mihindukulasooriya. 2022.
\newblock \href {https://doi.org/10.18653/v1/2022.trustnlp-1.7} {An empirical
  study on pseudo-log-likelihood bias measures for masked language models using
  paraphrased sentences}.
\newblock In \emph{Proceedings of the 2nd Workshop on Trustworthy Natural
  Language Processing (TrustNLP 2022)}, pages 74--79.

\bibitem[{Lal et~al.(2021)Lal, Ma, Aflalo, Howard, Simoes, Korat, Pereg,
  Singer, and Wasserblat}]{lal-etal-2021-interpret}
Vasudev Lal, Arden Ma, Estelle Aflalo, Phillip Howard, Ana Simoes, Daniel
  Korat, Oren Pereg, Gadi Singer, and Moshe Wasserblat. 2021.
\newblock \href {https://doi.org/10.18653/v1/2021.eacl-demos.17}
  {{I}nterpre{T}: An interactive visualization tool for interpreting
  transformers}.
\newblock In \emph{Proceedings of the 16th Conference of the European Chapter
  of the Association for Computational Linguistics: System Demonstrations},
  pages 135--142, Online. Association for Computational Linguistics.

\bibitem[{Lewis et~al.(2020)Lewis, Liu, Goyal, Ghazvininejad, Mohamed, Levy,
  Stoyanov, and Zettlemoyer}]{lewis-etal-2020-bart}
Mike Lewis, Yinhan Liu, Naman Goyal, Marjan Ghazvininejad, Abdelrahman Mohamed,
  Omer Levy, Veselin Stoyanov, and Luke Zettlemoyer. 2020.
\newblock \href {https://doi.org/10.18653/v1/2020.acl-main.703} {{BART}:
  Denoising sequence-to-sequence pre-training for natural language generation,
  translation, and comprehension}.
\newblock In \emph{Proceedings of the 58th Annual Meeting of the Association
  for Computational Linguistics}, pages 7871--7880, Online. Association for
  Computational Linguistics.

\bibitem[{Li et~al.(2021)Li, Xiao, Wang, Jang, and Carenini}]{li-etal-2021-t3}
Raymond Li, Wen Xiao, Lanjun Wang, Hyeju Jang, and Giuseppe Carenini. 2021.
\newblock \href {https://doi.org/10.18653/v1/2021.emnlp-demo.26} {T3-vis:
  visual analytic for training and fine-tuning transformers in {NLP}}.
\newblock In \emph{Proceedings of the 2021 Conference on Empirical Methods in
  Natural Language Processing: System Demonstrations}, pages 220--230, Online
  and Punta Cana, Dominican Republic. Association for Computational
  Linguistics.

\bibitem[{Liang et~al.(2021)Liang, Wu, Morency, and Salakhutdinov}]{LiangWMS21}
Paul~Pu Liang, Chiyu Wu, Louis{-}Philippe Morency, and Ruslan Salakhutdinov.
  2021.
\newblock \href {http://proceedings.mlr.press/v139/liang21a.html} {Towards
  understanding and mitigating social biases in language models}.
\newblock In \emph{Proceedings of the 38th International Conference on Machine
  Learning, {ICML} 2021, 18-24 July 2021, Virtual Event}, volume 139 of
  \emph{Proceedings of Machine Learning Research}, pages 6565--6576. {PMLR}.

\bibitem[{Nadeem et~al.(2021)Nadeem, Bethke, and
  Reddy}]{nadeem-etal-2021-stereoset}
Moin Nadeem, Anna Bethke, and Siva Reddy. 2021.
\newblock \href {https://doi.org/10.18653/v1/2021.acl-long.416} {{S}tereo{S}et:
  Measuring stereotypical bias in pretrained language models}.
\newblock In \emph{Proceedings of the 59th Annual Meeting of the Association
  for Computational Linguistics and the 11th International Joint Conference on
  Natural Language Processing (Volume 1: Long Papers)}, pages 5356--5371,
  Online. Association for Computational Linguistics.

\bibitem[{Nangia et~al.(2020)Nangia, Vania, Bhalerao, and
  Bowman}]{nangia-etal-2020-crows}
Nikita Nangia, Clara Vania, Rasika Bhalerao, and Samuel~R. Bowman. 2020.
\newblock \href {https://doi.org/10.18653/v1/2020.emnlp-main.154}
  {{C}row{S}-pairs: A challenge dataset for measuring social biases in masked
  language models}.
\newblock In \emph{Proceedings of the 2020 Conference on Empirical Methods in
  Natural Language Processing (EMNLP)}, pages 1953--1967, Online. Association
  for Computational Linguistics.

\bibitem[{Park et~al.(2019)Park, Na, Jo, Shin, Yoo, Kwon, Zhao, Noh, Lee, and
  Choo}]{park2019sanvis}
Cheonbok Park, Inyoup Na, Yongjang Jo, Sungbok Shin, Jaehyo Yoo, Bum~Chul Kwon,
  Jian Zhao, Hyungjong Noh, Yeonsoo Lee, and Jaegul Choo. 2019.
\newblock \href {https://doi.org/10.1109/VISUAL.2019.8933677} {{SANVis}: Visual
  analytics for understanding self-attention networks}.
\newblock In \emph{IEEE Visualization Conference (VIS)}, pages 146--150. IEEE.

\bibitem[{Radford et~al.(2018)Radford, Narasimhan, Salimans, Sutskever
  et~al.}]{radford2018improving}
Alec Radford, Karthik Narasimhan, Tim Salimans, Ilya Sutskever, et~al. 2018.
\newblock Improving language understanding by generative pre-training.

\bibitem[{Raffel et~al.(2020)Raffel, Shazeer, Roberts, Lee, Narang, Matena,
  Zhou, Li, and Liu}]{raffel2019exploring}
Colin Raffel, Noam Shazeer, Adam Roberts, Katherine Lee, Sharan Narang, Michael
  Matena, Yanqi Zhou, Wei Li, and Peter~J. Liu. 2020.
\newblock \href {https://doi.org/10.5555/3455716.3455856} {Exploring the limits
  of transfer learning with a unified text-to-text transformer}.
\newblock \emph{Journal of Machine Learning Research}, 21(1):5485--5551.

\bibitem[{Rissaki et~al.(2022)Rissaki, Scarone, Liu, Pandey, Klein,
  Eliassi-Rad, and Borkin}]{rissaki2022biascope}
Agapi Rissaki, Bruno Scarone, David Liu, Aditeya Pandey, Brennan Klein, Tina
  Eliassi-Rad, and Michelle~A Borkin. 2022.
\newblock \href {https://doi.org/10.1109/VDS57266.2022.00008} {{BiaScope}:
  Visual unfairness diagnosis for graph embeddings}.
\newblock In \emph{IEEE Visualization in Data Science (VDS)}, pages 27--36.

\bibitem[{Rudinger et~al.(2018)Rudinger, Naradowsky, Leonard, and
  Van~Durme}]{rudinger-gender}
Rachel Rudinger, Jason Naradowsky, Brian Leonard, and Benjamin Van~Durme. 2018.
\newblock \href {https://doi.org/10.18653/v1/N18-2002} {Gender bias in
  coreference resolution}.
\newblock In \emph{Proceedings of the 2018 Conference of the North {A}merican
  Chapter of the Association for Computational Linguistics: Human Language
  Technologies, Volume 2 (Short Papers)}, pages 8--14, New Orleans, Louisiana.
  Association for Computational Linguistics.

\bibitem[{Salazar et~al.(2020)Salazar, Liang, Nguyen, and
  Kirchhoff}]{salazar-etal-2020-masked}
Julian Salazar, Davis Liang, Toan~Q. Nguyen, and Katrin Kirchhoff. 2020.
\newblock \href {https://doi.org/10.18653/v1/2020.acl-main.240} {Masked
  language model scoring}.
\newblock In \emph{Proceedings of the 58th Annual Meeting of the Association
  for Computational Linguistics}, pages 2699--2712, Online. Association for
  Computational Linguistics.

\bibitem[{Schneider(2005)}]{schneider2005psychology}
David~J Schneider. 2005.
\newblock \emph{The psychology of stereotyping}.
\newblock Guilford Press.

\bibitem[{Shah et~al.(2020)Shah, Schwartz, and
  Hovy}]{shah-etal-2020-predictive}
Deven~Santosh Shah, H.~Andrew Schwartz, and Dirk Hovy. 2020.
\newblock \href {https://doi.org/10.18653/v1/2020.acl-main.468} {Predictive
  biases in natural language processing models: A conceptual framework and
  overview}.
\newblock In \emph{Proceedings of the 58th Annual Meeting of the Association
  for Computational Linguistics}, pages 5248--5264, Online. Association for
  Computational Linguistics.

\bibitem[{Sun et~al.(2019)Sun, Gaut, Tang, Huang, ElSherief, Zhao, Mirza,
  Belding, Chang, and Wang}]{sun-etal-2019-mitigating}
Tony Sun, Andrew Gaut, Shirlyn Tang, Yuxin Huang, Mai ElSherief, Jieyu Zhao,
  Diba Mirza, Elizabeth Belding, Kai-Wei Chang, and William~Yang Wang. 2019.
\newblock \href {https://doi.org/10.18653/v1/P19-1159} {Mitigating gender bias
  in natural language processing: Literature review}.
\newblock In \emph{Proceedings of the 57th Annual Meeting of the Association
  for Computational Linguistics}, pages 1630--1640, Florence, Italy.
  Association for Computational Linguistics.

\bibitem[{Tenney et~al.(2020)Tenney, Wexler, Bastings, Bolukbasi, Coenen,
  Gehrmann, Jiang, Pushkarna, Radebaugh, Reif, and Yuan}]{tenney2020language}
Ian Tenney, James Wexler, Jasmijn Bastings, Tolga Bolukbasi, Andy Coenen,
  Sebastian Gehrmann, Ellen Jiang, Mahima Pushkarna, Carey Radebaugh, Emily
  Reif, and Ann Yuan. 2020.
\newblock \href {https://doi.org/10.18653/v1/2020.emnlp-demos.15} {{The
  language interpretability tool: Extensible, interactive visualizations and
  analysis for NLP models}}.
\newblock \emph{EMNLP 2020 Demos}, 15.

\bibitem[{Van~der Maaten and Hinton(2008)}]{van2008visualizing}
Laurens Van~der Maaten and Geoffrey Hinton. 2008.
\newblock \href {http://jmlr.org/papers/v9/vandermaaten08a.html} {Visualizing
  data using {t-SNE}}.
\newblock \emph{Journal of machine learning research}, 9(11).

\bibitem[{Vig(2019)}]{vig-2019-multiscale}
Jesse Vig. 2019.
\newblock \href {https://doi.org/10.18653/v1/P19-3007} {A multiscale
  visualization of attention in the transformer model}.
\newblock In \emph{Proceedings of the 57th Annual Meeting of the Association
  for Computational Linguistics: System Demonstrations}, pages 37--42,
  Florence, Italy. Association for Computational Linguistics.

\bibitem[{Wallace et~al.(2019)Wallace, Tuyls, Wang, Subramanian, Gardner, and
  Singh}]{wallace-etal-2019-allennlp}
Eric Wallace, Jens Tuyls, Junlin Wang, Sanjay Subramanian, Matt Gardner, and
  Sameer Singh. 2019.
\newblock \href {https://doi.org/10.18653/v1/D19-3002} {{A}llen{NLP} interpret:
  A framework for explaining predictions of {NLP} models}.
\newblock In \emph{Proceedings of the 2019 Conference on Empirical Methods in
  Natural Language Processing and the 9th International Joint Conference on
  Natural Language Processing (EMNLP-IJCNLP): System Demonstrations}, pages
  7--12, Hong Kong, China. Association for Computational Linguistics.

\bibitem[{Weidinger et~al.(2021)Weidinger, Mellor, Rauh, Griffin, Uesato,
  Huang, Cheng, Glaese, Balle, Kasirzadeh et~al.}]{weidinger2021ethical}
Laura Weidinger, John Mellor, Maribeth Rauh, Conor Griffin, Jonathan Uesato,
  Po-Sen Huang, Myra Cheng, Mia Glaese, Borja Balle, Atoosa Kasirzadeh, et~al.
  2021.
\newblock \href {https://doi.org/10.48550/arXiv.2112.04359} {Ethical and social
  risks of harm from language models}.
\newblock \emph{arXiv preprint arXiv:2112.04359}.

\bibitem[{Wexler et~al.(2019)Wexler, Pushkarna, Bolukbasi, Wattenberg,
  Vi{\'e}gas, and Wilson}]{wexler2019if}
James Wexler, Mahima Pushkarna, Tolga Bolukbasi, Martin Wattenberg, Fernanda
  Vi{\'e}gas, and Jimbo Wilson. 2019.
\newblock \href {https://doi.org/10.1109/TVCG.2019.2934619} {The what-if tool:
  Interactive probing of machine learning models}.
\newblock \emph{IEEE Transactions on Visualization and Computer Graphics},
  26(1):56--65.

\bibitem[{Yan et~al.(2020)Yan, Gu, Lin, and Rzeszotarski}]{yan2020silva}
Jing~Nathan Yan, Ziwei Gu, Hubert Lin, and Jeffrey~M Rzeszotarski. 2020.
\newblock \href {https://doi.org/10.1145/3313831.3376447} {{Silva}:
  Interactively assessing machine learning fairness using causality}.
\newblock In \emph{Proceedings of the ACM CHI Conference on Human Factors in
  Computing Systems}, pages 1--13.

\bibitem[{Zhang et~al.(2019)Zhang, Zhao, Saleh, and Liu}]{zhang2019pegasus}
Jingqing Zhang, Yao Zhao, Mohammad Saleh, and Peter~J. Liu. 2019.
\newblock \href {http://arxiv.org/abs/1912.08777} {{PEGASUS}: Pre-training with
  extracted gap-sentences for abstractive summarization}.

\bibitem[{Zhao et~al.(2018)Zhao, Wang, Yatskar, Ordonez, and
  Chang}]{zhao-gender}
Jieyu Zhao, Tianlu Wang, Mark Yatskar, Vicente Ordonez, and Kai-Wei Chang.
  2018.
\newblock \href {https://doi.org/10.18653/v1/N18-2003} {Gender bias in
  coreference resolution: Evaluation and debiasing methods}.
\newblock In \emph{Proceedings of the 2018 Conference of the North {A}merican
  Chapter of the Association for Computational Linguistics: Human Language
  Technologies, Volume 2 (Short Papers)}, pages 15--20, New Orleans, Louisiana.
  Association for Computational Linguistics.

\end{thebibliography}
\bibliographystyle{acl_natbib}




\end{document}